\documentclass[american,a4paper,runningheads,envcountsect]{llncs}

\usepackage{babel}
\usepackage[utf8]{inputenc}
\usepackage[T1]{fontenc}

\usepackage{amsmath,amsthm,amssymb}
\usepackage{mathtools}
\usepackage{mathrsfs}
\usepackage{fca}
\usepackage{dsfont} 

\usepackage[colorlinks,citecolor=blue,urlcolor=black,hidelinks,linktocpage]{hyperref}
\usepackage[all]{hypcap}
\usepackage{cleveref}
\let\cref\Cref

\newtheorem{thm}{Theorem}[section]

\theoremstyle{definition}

\newtheorem{defn}[thm]{Definition}

\theoremstyle{remark}

\renewcommand{\epsilon}{\varepsilon}
\renewcommand{\phi}{\varphi}
\newcommand{\defeq}{\mathrel{\mathop:}=} 

\DeclareMathOperator{\spt}{spt}
\DeclareMathOperator{\diam}{diam}

\usepackage{microtype}
\usepackage{csquotes}
\usepackage{booktabs}
\usepackage{todonotes}
\presetkeys{todonotes}{color=blue!5}{}
\usepackage{paralist}

\usepackage[backend=bibtex,style=numeric-comp,doi=false,isbn=false,%
url=false]{biblatex}
\newcommand\blfootnote[1]{%
  \begingroup
  \renewcommand\thefootnote{}\footnote{#1}%
  \addtocounter{footnote}{-1}%
  \endgroup
}

\begin{document}

\title{Intrinsic dimension and its application to association rules}
\author{Tom Hanika\inst{1} \and Friedrich Martin Schneider\inst{2,3} \and Gerd Stumme\inst{1}}

\date{\today}

\institute{%
  Knowledge \& Data Engineering Group,
  University of Kassel, Germany\\[0.5ex]
  \and
  Institute of Algebra, TU Dresden, 01062 Dresden, Germany\\[0.5ex]
  \and
  Departamento de Matem\'atica, UFSC, Trindade, Florian\'opolis, SC, 88.040-900, Brazil\\[0.5ex]
  \email{tom.hanika@cs.uni-kassel.de,
    stumme@cs.uni-kassel.de, martin.schneider@tu-dresden.de}
}
\maketitle

\blfootnote{Authors given in alphabetical order.
  No priority in authorship is implied.}

\begin{abstract}
  The curse of dimensionality in the realm of association rules is
  twofold. Firstly, we have the well known exponential increase in
  computational complexity with increasing item set size.  Secondly,
  there is a \emph{related curse} concerned with the distribution of
  (spare) data itself in high dimension. The former problem is often
  coped with by projection, i.e., feature selection, whereas the best
  known strategy for the latter is avoidance.  This work summarizes
  the first attempt to provide a computationally feasible method for
  measuring the extent of dimension curse present in a data set with
  respect to a particular class machine of learning procedures.  This
  recent development enables the application of various other methods
  from geometric analysis to be investigated and applied in machine
  learning procedures in the presence of high dimension.
\end{abstract}

\keywords{Association~Rules, Geometric~Analysis, Curse~of~Dimensionality}

\section{Introduction}
\label{sec:introduction}
The \emph{curse of dimensionality} in machine learning is a well known
common place to flag the frontier to difficulty. However, in fact
there are at least two peculiarities of this, i.e., the combinatorical
explosion in high dimension and the (often) complicated data
distribution in high dimension. Even though both are connected to some
extent the latter is the object of investigation of this work, which
we call from now on \emph{dimension curse}. This effect is closely
related to the mathematical \emph{phenomenon of concentration of
  measure}, which was discovered by V.~Milman~\cite{Milman1983} and is
also known as the \emph{Lévy property}. There are various works
linking both worlds with the most comprehensive
being~\cite{Pestov1,Pestov2} by V.~Pestov. His axiomatic approach led
to a potent definition of \emph{intrinsic dimension}, which is,
however, computationally infeasible. Building up on his ideas we
presented in~\cite{sbomb} an applicable setup, which we summarize in
the following. For this we recall crucial notions from~\cite{sbomb}
and show how data sets may be analyzed for dimension curse. We
conclude our work with an exemplary application for association rules.

\section{Observable Diameters and Dimension Function}
\label{section:concentration}
Our approach for measuring the dimension curse for association rules
is based on methods from geometric analysis. Hence, we need some
mathematical structure that is accessible from both sides, geometrical
methods as well as data representation. For this we developed the
following~\cref{defi:ds} in~\cite{sbomb}. But first, let us briefly
recall some necessary basic mathematical notions. We call a
topological space $X$ \emph{polish} if $X$ is separable and there is a
complete metric generating the topology of $X$. A set of functions
$F\subseteq\mathbb{R}^{X}$ is called \emph{(pointwise) equicontinuous}
if $\forall x\in X,\epsilon>0$ there exists a neighborhood $U$ of $x$
in $X$ with $\diam f(U)\leq\epsilon$ for all $f\in F$. We utilize
frequently the push-forward measure idea. Let $S,T$ be measurable
spaces where $\mu$ is a measure on $S$ and $f:S\to T$ is a measurable
map. The \emph{push-forward measure} of $\mu$ is defined by
$f_{*}(\mu)(B)\coloneqq \mu(f^{-1}(B))$ for every measurable set
$B\subseteq T$. On a more technical note, for a measurable space $S$
with some measure $\mu$ and some measurable $T\subseteq S$ we denote
by $\mu\! \!  \upharpoonright_{T}$ the measure on the induced measure
space $T$ defined by
$(\mu \!\!\upharpoonright_{T} )(B) \defeq \mu (B)$ for every
measurable $B\subseteq T$. Finally, for a finite non-empty set $S$ we
denote by by $\nu_{S}$ the normalized counting measure on $S$.

\begin{defn}[Data Structure {\cite{sbomb}}]\label{defi:ds} A \emph{data structure} is a triple ${D=(X,\mu,F)}$ consisting of a Polish space~$X$ together with a Borel probability measure $\mu$ on $X$ and an equicontinuous set $F$ of real-valued functions on $X$, where the elements of $F$ will be referred to as the \emph{features} of $D$. We call $D_{\mathrm{triv}} \defeq (\{\emptyset\}, \nu_{\{\emptyset\}}, \mathbb{R})$ the \emph{trivial} data structure. Given a data structure $D = (X,\mu,F)$, we let \begin{displaymath}
	\widehat{D} \, \defeq \left. \left(X,\mu, F \cup \left\{ f \in \mathbb{R}^{X} \right| f \text{ constant} \right\} \right) .
      \end{displaymath}
We  call two data structures $D_{i} = (X_{i},\mu_{i},F_{i})$ $(i \in \{ 0,1 \})$ \emph{isomorphic} and write $D_{0} \cong D_{1}$ if there exists a homeomorphism $\phi \colon \spt \mu_{0} \to \spt \mu_{1}$ such that $\phi_{\ast}\! \left(\mu_{0}\! \! \upharpoonright_{\spt \mu_{0}}\right) = \mu_{1}\! \! \upharpoonright_{\spt \mu_{1}}$ and $(F_{1}\vert_{\spt \mu_{1}}) \circ \phi = F_{0}\vert_{\spt \mu_{0}}$. \end{defn}

In~\cite{sbomb} we elaborated on this definition. In particular, we
introduced a pseudo metric on the collection of all data structures, a
variant of Gromov's observable
distance~\cite[Chapter~3$\tfrac{1}{2}$.H]{Gromov99}. We utilize this
as a tool for analyzing high dimensional data, as proposed by
Pestov~\cite{Pestov1,Pestov2}. However, we will refrain to introduce
the specifics of this pseudo metric and refer to it informally for the
rest of this work.

The goal now is to find a mathematical sound \emph{dimension function}
for the collection of all data structures. We may skip the necessary
propositions and proceeding mathematical notions and state the
two most important properties to expect from a dimension function
informally. First, a dimension function should reflect the presence of
the concentration phenomenon in a data structure. More precisely, a
dimension function shall diverge on a sequence of data structures if
and only if this sequence has the Lévy property. Second, if a sequence
of data structures concentrates (w.r.t. the pseudo metric) to a
particular data structure, the dimension function shall concentrate to
its value on this data structure as well.  For the further
complete axiomatization we refer the reader to~\cite{sbomb}.

For reasons of space, we may not address the various technical
mathematical challenges, preparations and connections to geometric
analysis. We rather jump to the main result of~\cite{sbomb}, a new
quantity for expressing the extent to which a data structure is prone
to the dimension curse: the \emph{intrinsic dimension}. For this we adapt
further the ideas from~\cite[Chapter~3$\tfrac{1}{2}$]{Gromov99} about
observable diameters.

\begin{defn}[Observable Diameter~\cite{sbomb}]\label{definition:observable.diameter}
  Let $\alpha \geq 0$. We call~\cref{partdiam} the
  \emph{$\alpha$-partial diameter} of a Borel probability measure
  $\nu$ on $\mathbb{R}$ and we call~\cref{obsdiam} the
  \emph{$\alpha$-observable diameter} of a data structure
  $D = (X,\mu,F)$.
  \begin{align}
    &\mathrm{PartDiam}(\nu,1-\alpha)
    \defeq \inf \{ \diam (B) \mid B \subseteq \mathbb{R} \text{ Borel}, \, \nu
    (B) \geq 1-\alpha \}\label{partdiam}\\
    &\mathrm{ObsDiam}(D;-\alpha) \defeq
    \sup \{ \mathrm{PartDiam}(f_{\ast}(\mu),1-\alpha) \mid f \in F \}\label{obsdiam}
  \end{align}
\end{defn}

We showed in~\cite{sbomb} that the observable diameter is invariant
under isomorphisms of data structures and it fulfills a continuity
property with respect to the earlier mentioned pseudo metric. Building
up on this definition we can state:
The map $\partial_{\Delta} \colon \mathcal{D} \to [1,\infty]$ defined
by~\cref{dimfunc} is a dimension function.
\begin{equation}
  \partial_{\Delta}(D) \defeq \frac{1}{\left(\int_{0}^{1} \mathrm{ObsDiam}(D;-\alpha) \wedge 1 \, \mathrm{d}\alpha\right)^{2} } \qquad (D \in \mathcal{D})\label{dimfunc}
\end{equation}

\section{Example Experiment and Applications}
\label{sec:applications}
\begin{figure}[t]
  \centering
  \includegraphics[width=0.29\textwidth]{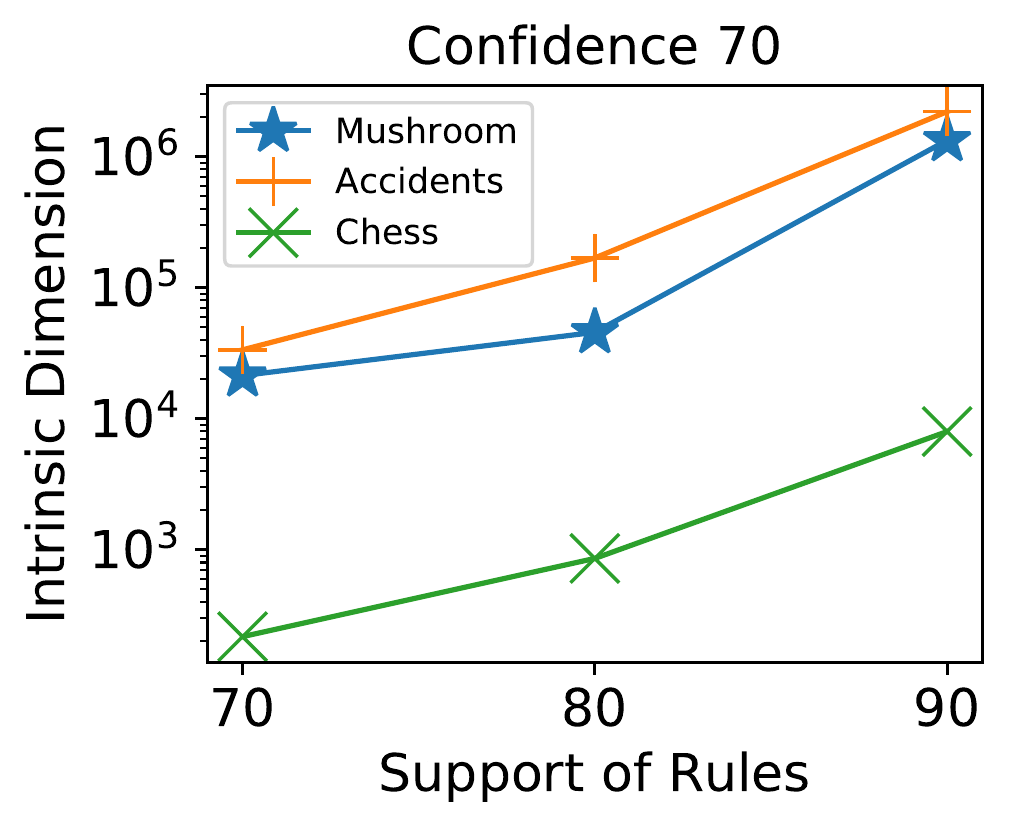}
  \includegraphics[width=0.29\textwidth]{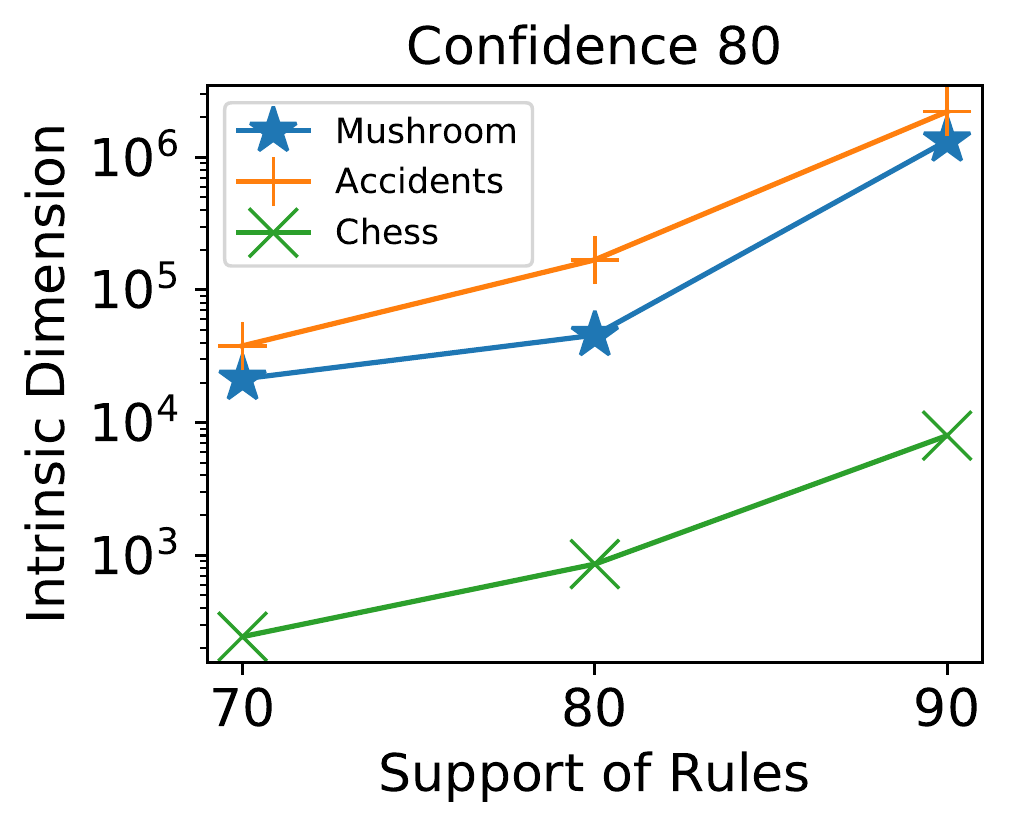}
  \includegraphics[width=0.29\textwidth]{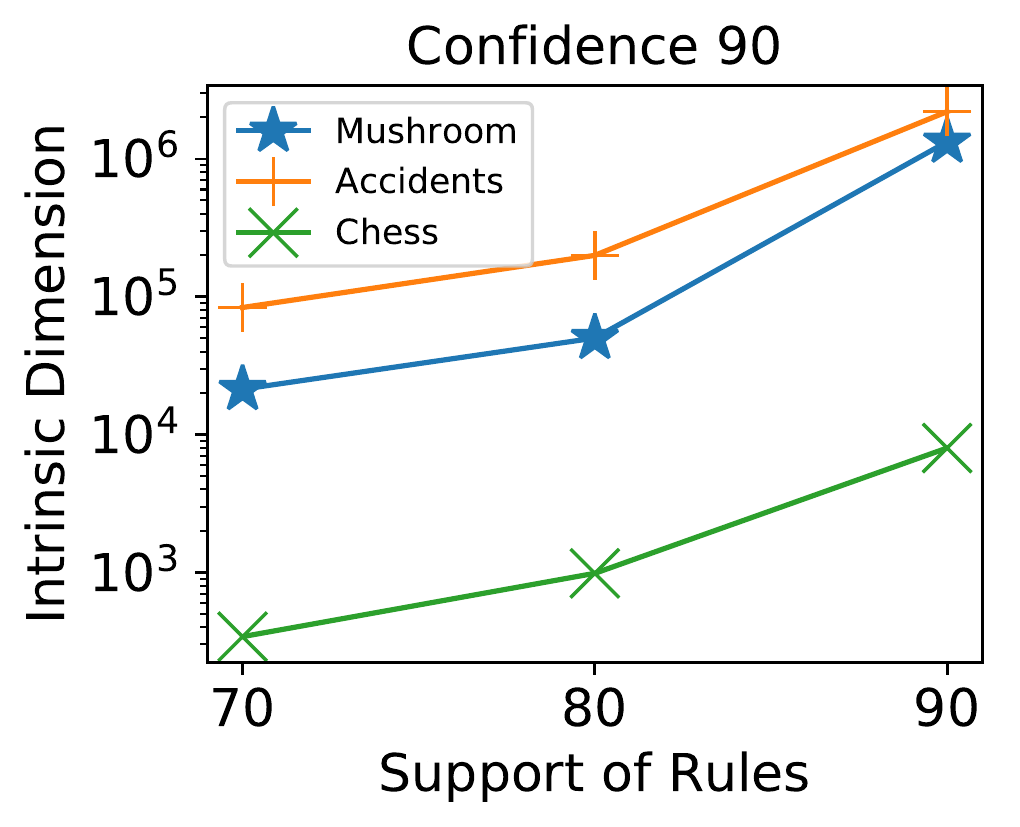}
  \caption{The intrinsic dimension for multiple data sets, supports
    and confidence.}
  \label{fig:plot}
\end{figure}

Distance functions, as often used in machine learning procedures, are
a natural candidate for feature functions. Hence, we might not need to
motivate the applicability of the intrinsic dimension for
those. However, the idea of dimension function in mathematical data
structures is able to cope with any kind of proper feature function
set. Therefore we decided for an exemplary application in association
rule mining. A possible adaption of data structures and observable
diameter could be done as follows: We consider a
set -- we restrict our example to non repeating transactions --
 of transactions $T=\{t_{1},\dotsc,t_{m}\}$ with
transactions $t_{i}\subseteq I$ where $I=\{i_{1},\dotsc,i_{n}\}$ is
called itemset. An \emph{assocation rule} on $I$ then is an element
$(X,Y)\in \mathcal{P}(I)\times \mathcal{P}(I)$ such that
$X\cap Y=\emptyset$ and
$\forall t\in T: X\subseteq t\Rightarrow Y\subseteq t$. We denote by
$R$ the set of all association rules for $T$, and by $X^{T}$ the
subset $S\subseteq T$ such that $\forall t\in S: X\subseteq t$. To
convert this data into a mathematical data structure like introduced
in~\cref{defi:ds} we take the following approach: $D=(T,\nu_{T},F(T))$
with
$F(T)\coloneqq\{\nu_{I}(Y)\cdot\mathds{1}_{X^{T}} \mid (X,Y)\in R\}$.
Hence, we consider the transactions as data points and the feature
functions are mappings from those data points to the support of a head
of a rule, i.e., the relative amount of items covered by this
particular rule. Using this setup the
$\mathrm{PartDiam}(\nu_{I}(Y)\cdot\mathds{1}_{X^{T}},1-\alpha)=
\nu_{I}(Y)$ if $\alpha<\nu_{T}(X^{T})<1-\alpha$ and $0$ otherwise, for
all rules $(X,Y)\in R$. Hence, this yields
$\mathrm{ObsDiam}(D,-\alpha)= \sup\{\nu_{I}(Y)\mid (X,Y)\in R,\alpha
<\nu_{T}(X^{T})<1-\alpha\}$.  We plotted in~\cref{fig:plot} multiple
example calculations for well known association rule minining data
sets, i.e., accident~\cite{accidentdata}, mushroom and
chess~\cite{UCI}.  We observe an increase in dimension with the
increase of support.

This is expected due to the antitone character of
feature sets. However, the slope differs
among the different data sets and confidence values, revealing the
ability of the particular feature sets to cover the data.

\subsubsection{Conclusion}
We presented in this work an mathematical approach for measuring the
dimension curse in machine learning. The novelty here is the
computationally feasible character. Besides the indicated application for association rules
there are various applications possible, and from the standpoint
of understanding the dimension curse, necessary. One particular
crucial application could be assessing the
results of dimension reduction procedures like Principle Component Analysis.\\

\noindent{\scriptsize
\textbf{Acknowledgments}\\
F.M.S.~acknowledges funding of the Excellence Initiative by the German Federal and State Governments, as well as the Brazilian CNPq, processo 150929/2017-0.
}

\sloppy
\printbibliography

\end{document}